\documentclass[10pt,twocolumn,letterpaper]{article}

\usepackage[pagenumbers]{iccv} 

\definecolor{iccvblue}{rgb}{0.21,0.49,0.74}
\usepackage[pagebackref,breaklinks,colorlinks,allcolors=iccvblue]{hyperref}

\def\paperID{12584} 
\def\confName{ICCV}
\def\confYear{2025}

\title{BridgeIV: Bridging Customized Image and Video Generation through Test-Time Autoregressive Identity Propagation}

\author{Panwen Hu$^{1}$, Jiehui Huang$^{2}$, Qiang Sun$^{1,3}$, Xiaodan Liang$^{1,2}$\thanks{Corresponding author} \\
$^{1}$Mohamed bin Zayed University of Artificial Intelligence \\
$^{2}$Sun Yat-sen University \quad $^{3}$University of Toronto\\
}

\begin{document}
\maketitle

\begin{abstract}
Both zero-shot and tuning-based customized text-to-image (CT2I) generation have made significant progress for storytelling content creation. In contrast, research on customized text-to-video (CT2V) generation remains relatively limited. Existing zero-shot CT2V methods suffer from poor generalization, while another line of work directly combining tuning-based T2I models with temporal motion modules often leads to the loss of structural and texture information. To bridge this gap, we propose an autoregressive structure and texture propagation module (STPM), which extracts key structural and texture features from the reference subject and injects them autoregressively into each video frame to enhance consistency. Additionally, we introduce a test-time reward optimization (TTRO) method to further refine fine-grained details. Quantitative and qualitative experiments validate the effectiveness of STPM and TTRO, demonstrating improvements of 7.8 and 13.1 in CLIP-I and DINO consistency metrics over the baseline, respectively.
\end{abstract}    
\vspace{-12pt}
\section{Introduction}
\label{sec:intro}

The field of video generation~\cite{he2022latent,blattmann2023align,wang2024lavie,yang2024cogvideox} has witnessed remarkable advancements in recent years, with significant breakthroughs in both quality and diversity of generated videos, owing to the enhancement of data quality and the optimization of algorithmic models. Given the inherent uncertainty in generative conditions, several studies have begun to explore conditional generation, such as utilizing image conditions to dictate the initial frame~\cite{blattmann2023stable,wang2023dreamvideo,xing2024dynamicrafter}, employing layout or contour conditions to govern structural composition~\cite{wang2024easycontrol,wang2025uniadapter}, and leveraging trajectories or camera poses to regulate motion dynamics~\cite{zheng2024cami2v,zhang2024tora,wang2025humanvid}. 

Recently, driven by practical application demands, the generation of long storytelling videos~\cite{hu2023reinforcement,hu2024storyagent,chen2024sitcom} has garnered increasing attention. Distinct from the aforementioned controllable generation approaches, this paradigm necessitates the consideration of content or subject control, thereby giving rise to personalized video generation. While personalized generation has achieved substantial progress in the realm of image synthesis~\cite{gal2022image,ruiz2023dreambooth,xiao2024fastcomposer}, studies in customized video generation remain relatively limited~\cite{she2025customvideox}. Consequently, this paper primarily focuses on the investigation of personalized generation within the video domain.

\begin{figure}[th!]
\centering

\includegraphics[width=0.45\textwidth]{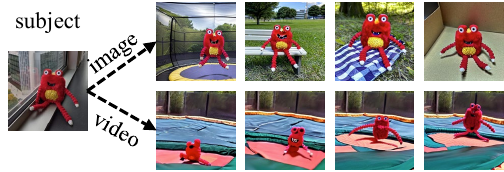} 
\caption{The first row presents the generated customized images, while the second row displays the video result. Although the customized image generation model achieves promising results, directly integrating a temporal motion module leads to structural and texture distortions in the video}
\vspace{-12pt}
\label{fig:problem definition}
\end{figure}

The objective of Customized Text-to-Video (CT2V) generation is to generate videos that simultaneously satisfy the given textual descriptions and maintain consistency with the reference subject, using text prompts and reference images as inputs. Current CT2V approaches can be broadly categorized into two types: zero-shot methods and fine-tuning methods. Zero-shot methods~\cite{he2024id,jiang2024videobooth} typically design a generalizable subject feature extractor to capture the identity features, which are then injected into the video generation process. While this approach offers advantages in inference efficiency, zero-shot methods often struggle with generating high-fidelity results for uncommon subjects not seen in the training data. Fine-tuning methods~\cite{wei2024dreamvideo} typically follow a Customized Text-to-Image (CT2I) generation strategy, where a model is fine-tuned on several subject images and subsequently inflated to the video domain~\cite{chefer2024still}. However, as shown in Figure~\ref{fig:problem definition}, directly adapting CT2I models to video generation with the pretrained temporal motion module often results in reduced subject consistency, with the subject’s structure and texture undergoing unintended deformations as the video evolves. 

To bridge the gap between the CT2I model and video generation, we believe that extracting, propagating, and enhancing the subject's structure and texture are crucial for improving video consistency. Therefore, this paper proposes a framework to solve the challenge from these three aspects. Specifically, our method consists of three stages: the customization stage, the structure and texture injection stage, and the latent enhancement stage. In the customization stage, inspired by mainstream CT2I approaches~\cite{pang2024attndreambooth,nam2024dreammatcher}, we adopt texture inversion~\cite{gal2022image} and DreamBooth~\cite{ruiz2023dreambooth} techniques to customize a T2I model. In the structure and texture injection stage, we integrate the CT2I model with a pre-trained temporal motion module to generate an initial video. During this process, we propose an autoregressive Structure and Texture Propagation Module (STPM), which consists of a Structure Propagation Module (SPM) and a Texture Propagation Module (TPM). These modules extract the subject’s structure and texture information from the CT2I model and inject them into video frames, ensuring structural and texture consistency throughout the video. In the latent enhancement stage, we propose a Test-Time Reward Optimization (TTRO) method, which iteratively refines the latent distribution of the video by leveraging feature similarity comparison. This approach further enhances the consistency of fine-grained details in the generated subject, improving overall visual coherence in the final video.


We evaluate our proposed model on the publicly available dataset~\cite{hao2023vico} and compare it with existing CT2V methods. Both qualitative and quantitative results demonstrate that our approach achieves state-of-the-art performance in maintaining subject identity consistency, outperforming the baseline with improvements of 7.8 and 13.1 in CLIP-I and DINO consistency metrics, respectively. In summary, our main contributions are as follows:
\begin{enumerate}
    \item We present a novel three-stage CT2V framework and introduce an innovative Structure and Texture Propagation Module (STPM) that effectively bridges the gap between the CT2I model and video generation.
    \item To enhance video quality, we introduce the concept of test-time scaling into CT2V and propose a test-time reward optimization (TTRO) method, which refines the generated results in terms of consistency.
    \item Finally, we implement existing CT2V methods and conduct qualitative, quantitative, and ablation experiments on the public benchmark to demonstrate the effectiveness and superiority of the proposed method.
\end{enumerate}

\section{Related work}
\label{sec:related work}
Customized text-to-video (CT2V) generation aims to produce video content that closely aligns with a given reference image and textual description by integrating both inputs. With the remarkable advancements in customized text-to-image (CT2I) generation~~\cite{hao2023vico,chen2023photoverse,han2023svdiff,liu2023cones}, CT2V has garnered increasing attention. Existing CT2V methods can be categorized into three main approaches: zero-shot approaches, tuning-based methods—both of which have been introduced in CT2I—and a two-stage approach. The two-stage approach~\cite{zhuang2024vlogger,hu2024storyagent} typically consists of a CT2I model followed by an image-to-video (I2V) generation model~\cite{xing2024dynamicrafter,blattmann2023stable,ren2024consisti2v}. For example, VideoStudio~\cite{long2024videostudio} first trains a zero-shot CT2I model to generate customized images and subsequently employs a dedicated I2V model to animate the generated images into videos. Similarly, any existing CT2I model~~\cite{hao2023vico,chen2023photoverse,han2023svdiff,liu2023cones} can be combined with an I2V model to form a two-stage pipeline. However, the limited generalization capability of I2V models often leads to unexpected subject deformations during the animation process.

To enhance the consistency of customized subjects in videos, zero-shot approaches~\cite{he2024id,jiang2024videobooth} adopt a single-stage framework that directly generates videos from a reference image and a text prompt. Methods in this category, such as VideoBooth~\cite{jiang2024videobooth}, typically collect large-scale datasets to train a general identity feature extractor. The extracted features are then injected into the backbone network through specialized mechanisms, and the feature extractor and the backbone network will be trained in an end-to-end manner. While these approaches offer significant advantages in generation efficiency, they come with high training costs and limited generalization ability. As a result, their performance deteriorates when dealing with uncommon subjects. 

Another category is the tuning-based approach~\cite{wu2024customcrafter,chefer2024still,wang2024customvideo,li2024personalvideo,chen2023videodreamer}, which we adopt in our method and currently demonstrates superior fidelity in CT2V. Some methods, such as DreamVideo~\cite{wei2024dreamvideo}, directly train a token embedding and an adapter on a video model using customized images. However, training a video model with image data often leads to overfitting, causing the loss of motion information. Another class of methods, such as Magic-Me~\cite{ma2024magic}, first trains a CT2I model and then integrates it with a pre-trained motion module, enabling direct video generation. However, the gap between CT2I model and video generation leads to subject deformation and texture loss in the generated videos. To bridge this gap, we propose an autoregressive Structure and Texture Propagation Module (STPM) that injects and propagates the structure and texture information of the reference subject across all video frames in an autoregressive manner. Furthermore, to enhance fine-grained detail consistency, we introduce a test-time reward optimization method to refine the latent distribution of the generated videos.

\section{Preliminary}
\subsection{Latent Diffusion Model}
Our method is built upon a spatiotemporal decoupled text-to-video (T2V) model, AnimateDiff~\cite{guo2023animatediff}, which is inflated from the text-to-image (T2I) model, Stable Diffusion~\cite{rombach2021highresolution}, by incorporating temporal motion module behind the original spatial blocks. Consequently, our approach follows the conventional training and generation pipeline of latent diffusion models (LDMs)~\cite{rombach2021highresolution}.

Specifically, during the first stage, i.e., the customization stage, we finetune a text embedding and the model parameters on the T2I model using the text inversion~\cite{gal2022image} and Dreambooth~\cite{ruiz2023dreambooth} methods. A subject image $I$ is first encoded into a latent representation $z_0$ using an image encoder. This latent representation $z_0$ is then corrupted with Gaussian noise $\epsilon$ and noise level $t$ through a forward diffusion process, forming a noisy latent representation $z_t$. The diffusion model subsequently takes the noisy latent representation $z_t$ and a text condition $c$ containing a special subject token $<S^*>$ as input, aiming to predict the noise $\epsilon$ added to the original latent representation.

Thus, the training objective of the diffusion model is formulated as:
\begin{equation}
\label{eqn:diffusion loss}
    \mathbb{E}_{z_0,c,\epsilon,t}[\|\epsilon_{\theta}(z_t, c, t)-\epsilon \|]
\end{equation}
where $\epsilon_{\theta}$ represents the model prediction. During inference, a random Gaussian noise $\hat{z}_T$ is sampled and fed into the diffusion model, which predicts the noise component. This predicted noise is then utilized to iteratively denoise $\hat{z}_T$ through the reverse diffusion process, progressively refining the latent representation for $t= T,\cdots,1$ to get the hidden representation $\hat{z}_0$ of the sampled image. This process is formulated as:
\begin{equation}
\label{eqn:denoising}
    \hat{z}_{t-1} = \frac{1}{\sqrt{\alpha_t}}(\hat{z}_t - \frac{1-\alpha_t}{\sqrt{1-\bar{\alpha}}}\epsilon_{\theta}(\hat{z}_t,c,t)) + \sigma_t\epsilon,
\end{equation}
where $\sigma_t = \frac{1-\bar{\alpha}_{t-1}}{1-\bar{\alpha}_t}\beta_t$,$\beta_t = 1-\alpha_t$.

\subsection{Cross-attention and Self-attention}
Our base model, AnimateDiff~\cite{guo2023animatediff}, employs a UNet architecture that includes cross-attention modules, self-attention modules, and motion modules. Motion modules process features along the temporal dimensions, while the injection of conditional information, such as text prompts, and the feature aggregation primarily rely on cross-attention and self-attention modules~\cite{song2020denoising}, respectively. In the cross-attention module, the hidden latents are projected as queries $Q$, while the condition embeddings are projected as keys $K$ and values $V$. The latents are then updated through the cross-attention mechanism:
\begin{equation*}
    CA(Q,K,V)=Softmax(\frac{QK^{T}}{\sqrt{d}})V
\end{equation*}
Here, $d$ refers to the channel dimensions, $Softmax()$ is applied over the keys for each query so that the condition information encoded in the projected value matrix $V$ is selectively injected into hidden latents. Meanwhile, by inspecting the attention map $A=KQ^{T}/\sqrt{d}$, we can identify the regions in the hidden latents that correspond to each condition token, such as the special reference subject token $<S^*>$. Thus, analyzing the distribution of the attention map associated with
$<S^*>$ allows us to assess the structure information of the reference subject~\cite{balaji2022ediff,hertz2022prompt}. This insight can be leveraged to optimize the hidden latents to enhance subject fidelity.

On the other hand, the self-attention module employs an attention strategy similar to that of the cross-attention module to aggregate features. In the self-attention module, the queries, keys, and values are all derived from hidden features, enabling a high degree of editability. As analyzed in previous works~\cite{tumanyan2023plug,tewel2023key}, the self-attention module controls the appearance attributes, such as color and texture, through the values $V$. 
\section{Method}
\begin{figure*}[ht!]
\centering
\includegraphics[width=1\textwidth]{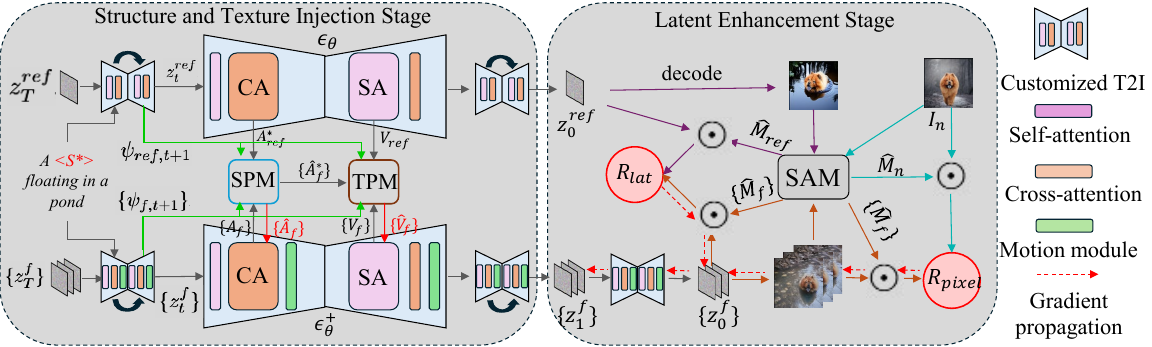} 
\caption{The processes of the structure and texture injection stage (the second stage) and the latent enhancement stage (the third stage). In the second stage, SPM extracts the reference subject's structure information $A^{*}_{ref}$ from the cross-attention of the CT2I model and propagates it to video frames through $\{\hat{A}_f\}$, using matching flows computed from the features, $\psi_{ref,t+1}$ and $\{\psi_{f,t+1}\}$ of the previous denoising step. Similarly, TPM utilizes the structure information $\{\hat{A}^*_{f}\}$ and matching flow to transfer the reference subject's texture information $V_{ref}$ into the video. In the third stage, two reward functions are designed in the latent and pixel domains, respectively, to correct the latent distribution, enhancing the consistency of the foreground.}
\vspace{-12pt}
\label{fig:framework}
\end{figure*}

Given a set of $N$ reference images $\mathbf{I}=\{I_n\}_{1}^{N}$ and a target text prompt $c$ containing a special token $<S^*>$, the objective of customized text-to-video (CT2V) generation is to generate a video $V=\{F_f\}_{f=1:J}$ that faithfully incorporates the reference subject $<S^*>$ while adhering to the given text prompt $c$. $J$ is the number of video frames $F_f$. Training-based (zero-shot) methods typically generate videos by conditioning on one or more reference images. However, these approaches rely on large-scale annotated datasets for training and exhibit poor generalization to unseen objects. Another line of zero-shot approaches~\cite{he2024id,jiang2024videobooth} leverage advanced customized text-to-image (CT2I) generation methods~\cite{hao2023vico,chen2023photoverse} first to synthesize subject-specific images, which are then animated using an off-the-shelf image-to-video (I2V) model~\cite{xing2024dynamicrafter,blattmann2023stable}. However, the effectiveness of this two-stage pipeline is heavily dependent on the I2V model’s capability, and current I2V models struggle with animating rare or complex objects.

To improve generalization, recent methods explore tuning-based solutions, where a customized video generation model is fine-tuned using only a few reference images $\mathbf{I}$. Our method follows this paradigm but differs in several key aspects. Existing approaches~\cite{wei2024dreamvideo} either fine-tune the entire video model using images—leading to the loss of motion dynamics—or only fine-tune the embedding of a special token $<S^*>$~\cite{ma2024magic}, which limits the model’s ability to accurately represent the subject. To address this issue, we propose a three-stage framework for CT2V, including the customization stage, the structure and texture injection stage, and the latent enhancement stage. At the customization stage, inspired by CT2I generation, we employ text inversion~\cite{gal2022image} and Dreambooth~\cite{ruiz2023dreambooth} techniques to train a customized T2I model $\epsilon_{\theta}$. This process learns the embeddings of the special token $<S^*>$ and the customized weights of the T2I model $\epsilon_{\theta}$. 

Although the T2I model $\epsilon_{\theta}$ at this stage can generate high-quality customized images, the modality gap between image and video domains remains a challenge. Directly adding pre-trained temporal motion modules to $\epsilon_{\theta}$ to form the video model $\epsilon_{\theta}^{+}$ alters the subject’s structural and texture distributions, leading to a decline in subject consistency across generated video frames~\cite{chefer2024still}. To bridge this modality gap, as shown in Figure~\ref{fig:framework}, we propose a novel autoregressive Structure and Texture Propagation Module (STPM), which includes a Structure Propagation Module (SPM) and a Texture Propagation Module (TPM), in the structure and texture injection stage. This module maximally transfers the structural and texture information from the generated reference images to every frame during video generation, while maintaining the smoothness of the video.

Due to the potential loss of fine-grained subject details (e.g., local colors) between the generated image and the original reference images $\textbf{I}$, the texture information extracted from the generated image may sometimes be insufficient to ensure that the subject in the video aligns perfectly with the subject in $\textbf{I}$. To address this issue, we propose a Test-Time Reward Optimization (TTRP) method in the third stage, which further refines the video latent representation, ensuring that the generated subject continuously approaches the true subject.

\subsection{Structure and Texture Propagation}
\begin{figure}[ht!]
\centering
\includegraphics[width=0.48\textwidth]{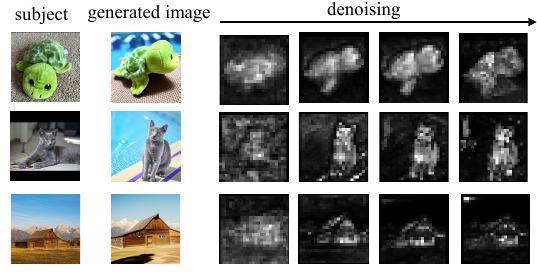} 
\caption{The visualization of normalized attention maps of the special token in different denoising steps. The highlighted regions in the attention maps correspond to the generated reference image (second column), indicating that the attention maps clearly capture the subject's structural information. }
\vspace{-12pt}
\label{fig:cross-attention map}
\end{figure}

\noindent \textbf{Structure Propagation} 
To enhance the consistency between the video subject structure and the reference subject structure, we need to explore two key aspects: how to extract structural information and how to inject structural information. From the cross-attention map shown in Figure~\ref{fig:cross-attention map}, we observe that in the T2I model $\epsilon_{\theta}$, the cross-attention modules play a significant role in shaping the subject's structure. Therefore, we extract the attention map corresponding to the special token $<S^*>$ as the structural information and autoregressively inject it into each video frame through a semantic matching flow $o^{F_i \rightarrow F_j}$, which is a displacement field~\cite{cho2021cats,hong2022neural,truong2023pdc} used to warp the subject in $F_i$ to match the subject in $F_j$. 

When extracting structural information, we enhance structural diversity by extracting it from the generated image $F_{ref}$. Specifially, let $i^*$ denote the position of $<S^{*}>$ in the target prompt $c$, for each denoising timestep $t$, the corresponding subject structure distribution $A^{*}_{ref,l} \in \mathbb{R}^{h_l*w_l \times 1}$ can be obtained from the cross-attention map $A_{ref, l}=Q_{ref,l}K^{T}_{ref,l}/\sqrt{d}, A_{ref, l} \in \mathbb{R}^{h_l*w_l \times L}$ of $\epsilon_{\theta}$ as $A^{*}_{ref,l}=A_{ref,l}[:,i^*]$, where  $l$, $L$ represent the layer number of the attention map and the length of target prompt tokens, respectively, and $h_l$, $w_l$ represent the height and the width of $Q_{ref, l}$, respectively. For simplicity, we will omit the layer notation $l$ in the following mathematical formulas. Then the target structure $\{A^{*}_f\}_{f=1:J}$ by autoregressive warping using the matching flow $\{o^{F_i \rightarrow F_j}\}$,
\begin{equation}
    A^{*}_f = \begin{cases}
        Warp(A_{ref}^{*}, o^{F_{ref}\rightarrow F_{f}}), & \text{if} f = 1, \\
        Warp(A_{f}^{*}, o^{F_{f}\rightarrow F_{f+1}}), & \text{otherwise.}\\
    \end{cases}
\end{equation}
Consequently, the final cross-attention maps $\{A_f\}_{f=1:J}, A_f \in \mathbb{R}^{h*w \times L}$ for the video $V$ in $\epsilon_{\theta}^{+}$ are obtained by concatenation operation $concat()$,
\begin{equation}
    \hat{A}_f = concat(A_f[:,1:i^{*}], A^{*}_f, A_f[:,i^*:]).
\end{equation}
Thus, the output of the cross attention blocks in $\epsilon_{\theta}^{+}$ becomes $\{Softmax(\hat{A}_f)V_f\}_{f=1:J}$. Through the above process, the structural information can be seamlessly and smoothly injected into the video frames.

\noindent \textbf{Matching Flow} Matching flow is the key to successfully injecting structural information. In our framework, we adopt a standard semantic matching workflow~\cite {hong2022neural,truong2023pdc} to compute the matching flow between two image frames. These conventional pipelines rely on a pretrained feature extractor to extract features from image pairs and subsequently compute their correspondence. However, finding good features under the diffusion framework is not trivial due to the noisy nature of video frames and reference images in the reverse diffusion process.

To address this issue, inspired by previous works~\cite{tang2023emergent,zhang2023tale}, we focus on the features from the second-to-last decoder layer of the denoising U-Net $\epsilon_{\theta}$ (and $\epsilon_{\theta}^{+}$), considering its spatial size and the richness of semantic information. Feature maps with excessively high resolution (from higher decoder layers) lead to excessive computational costs, while those with overly low resolution contain insufficient semantic information. let $\psi_{f, t+1} \in \mathbb{R}^{h\times w \times d}$ denote the features of frame $F_f$ at denosing timestep $t+1$, then the matching cost $C_{t+1} \in \mathbb{R}^{(h\times w) \times (h \times w)}$ between the frame pair is computed as the cosine similarity between the feature vectors of all positions, which is formulated as:
\begin{equation}
    C_{t+1}^{f \rightarrow f+1}[i,j]= \frac{\psi_{f,t+1}(i)\cdot \psi_{f+1,t+1}(j)}{\|\psi_{f,t+1}(i)\| \|\psi_{f+1,t+1}(j)\|},
\end{equation}
where $i, j \in [0,h) \times [0, w)$, and $\|\|$ denotes $l-2$ normalization. Subsequently, we derive the matching flow $o_t^{F_f->F_{f+1}} \in \mathbb{R}^{h \times w \times 2}$ used at time step $t$ by applying argmax operation~\cite{cho2021cats} on  $C_{t+1}^{f \rightarrow f+1}$. It is worth noting that when computing the matching flow for attention maps of different resolutions, we interpolate the features to the corresponding resolution before calculating the matching cost.

\begin{figure}[ht!]
\centering
\includegraphics[width=0.46\textwidth]{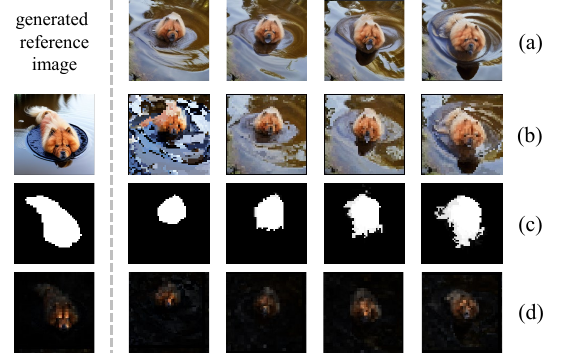} 
\caption{The visualization of the warping results with the matching flows. The first column represents the generated reference image, its foreground mask, and the results after multiplying with the attention map. (a) The corresponding generated video frames. (b) The warping results from the previous frame. (c) The warped foreground masks. (d) The warping results multiplied with the attention maps. We can observe that the subject from the reference image or video frame is correctly warped to the appropriate position in the next frame, providing the subject's texture information.}
\vspace{-12pt}
\label{fig:teaser}
\end{figure}

\noindent \textbf{Texture Propagation} In addition to structural information, the texture details of the subject may also be lost during the transition from the image modality to the video modality. To address this, we propose an autoregressive texture propagation module (TPM) to propagate the reference subject’s texture information across video frames. Previous research~\cite{balaji2022ediff} has shown that self-attention module plays a crucial role in integrating texture information stored in the value variable. Therefore, our TPM enhances the values $\{V_f\}_{f=1:J}, V_f \in \mathbb{R}^{(h*w) \times d}$ of the self-attention module. Specifically, we first inject the reference texture into the first frame of the video using the matching flow $\{o\}$ introduced above. Then, the texture is gradually propagated to the subsequent frames of the video. Thus, the enhanced values $\hat{V}_f$ for each frame is computed as:
\begin{equation}
    \hat{V}_f = \begin{cases}
        Warp(V_{ref}, o^{F_{ref} \rightarrow F_{f}}), & \text{if} f=1, \\
        Warp(V_{f}, o^{F_{f-1} \rightarrow} F_{f}), & \text{otherwise}.
    \end{cases}
\end{equation}
However, a straightforward warping operation may introduce inconsistencies in the appearance of elements outside the subject, such as background variations. Therefore, it is crucial to effectively isolate the subject from the background to ensure that texture propagation affects only the subject region.

To achieve this, we leverage the cross-attention map $\hat{A}_f^*$ of the special token as a soft foreground mask to filter out background information. Specifically, let $\mathbf{M}_f$ denote the soft foreground mask computed at frame $f$, which can be derived by applying a minmax norm on the cross-attention map $\hat{A}^*_f$, then we employ this mask to selectively blend the propagated texture features, ensuring that only the subject region is influenced while preserving background in the original frame. The value $\hat{V}_f$ is updated as :

\begin{gather*}
    \hat{V}_f = Warp(V_{f-1}, o^{F_{f-1} \rightarrow F_{f}}) \odot M_f + V_f \odot (1-M_f) \\
    M_f = norm(\hat{A}^*_{f}),
\end{gather*}
where $\odot$ represents the Hadamard product~\cite{horn1990hadamard}. Subsequently, the updated values will replace the original values to complete the self-attention computation $Softmax(Q_fK_f^T/\sqrt{d})\hat{V}_f$. It is worth noting that, as is typically the case in U-Net, the self-attention module appears before the cross-attention module in each block. Therefore, the cross-attention map used by the self-attention module comes from the previous cross-attention module, whereas the first self-attention module retains the original computation unchanged.

\subsection{Test-time Reward Optimization}
The STPM is performed at each denoising step. However, in the early time steps, the attention map and the values in the self-attention and cross-attention modules are noisy and lack fine-grained subject details. As a result, in the generated video, there still exists a certain discrepancy in the fine-grained details of the object compared to the original reference image. Therefore, in the third stage of our method, to correct this discrepancy, we propose a Test-Time Reward Optimization (TTRO) approach to refine the distribution of the latent representation to maximize the designed reward functions. 

Let $\{z_{f, 0}\}$ denote the latent representation obtained from the DDIM denoising~\cite{song2020denoising} in the second stage. Before the correction, we perform a diffusion forward process to obtain the noisy latent $\{z_{f, 1}\}$, which is then fed into the model $\epsilon_{\theta}^{+}$ to predict the noise $\epsilon$, and the new latents $\{\hat{z}_{f, 0}\}$ are computed with Eq.\ref{eqn:denoising}. Similary, the new latents $\hat{z}_{ref, 0}$ of the generated reference image can be obtained in the same manner. Intuitively, the subject region in the new latent $\hat{z}_{f,0}$ is expected to closely match the subject region of the reference latent $\hat{z}_{ref,0}$. Therefore, we propose using the cosine similarity of the average latents in their respective regions as the reward function, where the subject region $\hat{M}_f$ is identified by applying the segmentation method grounding-SAM~\cite{ren2024grounded} on the images decoded from the latents. The latent reward function can be written as:
\begin{equation*}
    R_{lat} = \sum_{f=1}^{J} \frac{mean(\hat{z}_{f,0} \odot \hat{M}_f)\cdot mean(\hat{z}_{ref,0} \odot \hat{M}_{ref})}{\|mean(\hat{z}_{f,0} \odot \hat{M}_f)\|\|mean(\hat{z}_{ref,0} \odot \hat{M}_{ref})\|},
\end{equation*}
where $mean()$ denotes the average function along the spatial dimension, and only the foreground region, i.e., $\hat{M} = 1$, is counted.

In addition to the latent domain, we further measure the reward in the pixel domain by using the off-the-shelf CLIP model~\cite{radford2021learning} $Clip()$ to optimize the latents. We can obtain the frames $\{F_f\}$ in pixel domain by decoding the latents $\{\hat{z}_{f, 0}\}$ with the pretrained VAE decoder~\cite{he2022latent}. Then, we measure the clip score between the generated frames and the original customized images $I_n$, and the reward function in pixel domain is formulated as:
\begin{equation*}
    R_{pixl} = \frac{1}{JN}\sum_f\sum_n Clip(F_f \odot \hat{M}_{f}, I_n \odot \hat{M}_n).
\end{equation*}
Finally, the noisy latent $\hat{z}_{f,1}$ is updated as follow:
\begin{equation*}
    \hat{z}_{f,1} = \hat{z}_{f,1} + \lambda \nabla_{\hat{z}_{f,t}}(R_{lat} + R_{pixel})
\end{equation*}
By iterating this process continuously, the reward value gradually increases, indicating that the subject in the video inferred from $\{\hat{z}_{f,1}\}$ becomes increasingly similar to the reference subject.

\section{Experiment}

\begin{figure*}[ht!]
\centering
\includegraphics[width=1\textwidth]{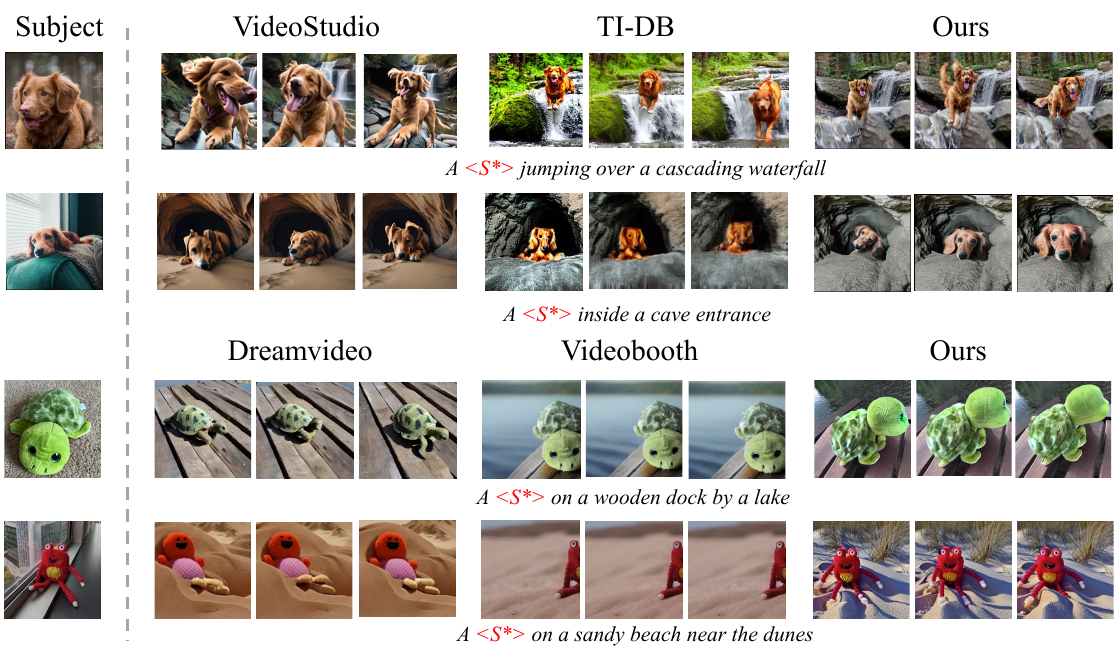} 
\vspace{-24pt}
\caption{The comparisons with different state-of-the-art CT2V methods, including two-stages methods, VideoStudio~\cite{long2024videostudio} and the baseline TI-DB, tuning-based method, Dreamvideo~\cite{wei2024dreamvideo}, and zero-shot method, Videobooth~\cite{jiang2024videobooth}. It can be observed that the two-stage method is prone to structural deformation, while traditional tuning-based and zero-shot methods have weak representation capabilities for unique objects, resulting in poor consistency in the generated results. }
\vspace{-12pt}
\label{fig:sota comparison}
\end{figure*}

\subsection{Experimental Settings}
\noindent \textbf{Dataset.} To validate the effectiveness of the proposed method, we conducted experiments using a dataset, Vico~\cite{hao2023vico}, for evaluating the customized T2I task. Vico dataset was collected from previous datasets and consists of 16 customized subjects, each containing 4 to 8 images. Additionally, we employed a more challenging set of prompts provided by~\cite{nam2024dreammatcher}, which includes 24 prompts featuring large displacement, occlusion, and novel-view synthesis. In our experiments, we generated 4 videos for each prompt, resulting in a total of 1,536 videos for evaluation.

\noindent \textbf{Comparion Methods.} In this experiment, we categorize the baseline methods into three groups: two-stage methods, zero-shot methods, and tuning-based methods. The two-stage methods include VideoStudio~\cite{long2024videostudio} and two additional approaches that first generate images using a customized T2I model and subsequently synthesize videos using an I2V model, specifically CogVideoX. For this category, we employ text inversion + DreamBooth (TI-DB) and MS-Diffusion~\cite{wang2024ms} as the T2I models. The zero-shot methods include VideoBooth~\cite{jiang2024videobooth} and SSR-Encoder~\cite{zhang2024ssr}, where SSR-Encoder was originally designed for the customized T2I task but can be adapted for video generation by integrating Animatediff’s motion module. Additionally, we compare our approach with Magic-Me~\cite{ma2024magic} and DreamVideo~\cite{wei2024dreamvideo}, which adopt a similar tuning-based strategy. By evaluating our method against these diverse approaches, we aim to comprehensively demonstrate its effectiveness in customized video generation.

\noindent \textbf{Evaluation Metrics.} To evaluate the subject customization capability of our method, we adopt three evaluation metrics following prior work~\cite{wei2024dreamvideo,wang2024customvideo}. CLIP-T~\cite{radford2021learning} measures text-image alignment by computing the average cosine similarity between the CLIP image embeddings of all generated frames and the corresponding text embedding. CLIP-I assesses visual similarity by calculating the average cosine similarity between the CLIP image embeddings of all generated frames and the embeddings of the original subject images. Additionally, DINO-I~\cite{ruiz2023dreambooth} serves as an alternative metric for visual similarity, where we utilize ViT-S/16 DINO~\cite{zhang2022dino} to extract embeddings from all generated frames and reference images and compute their cosine similarity. These metrics collectively provide a comprehensive evaluation of both semantic alignment and visual fidelity in the generated customized videos.

\begin{table}[ht!]
\vspace{-12pt}
\centering
\begin{tabular*}{\hsize}{@{\extracolsep{\fill}}c|c|ccc}
\hline

 Method & Type &  CLIP-T $\uparrow$	& CLIP-I $\uparrow$	& DINO $\uparrow$  \\
\hline
TI-DB	& T. &29.8	& 72.2	& 49.5	 \\
MS-Diffusion& T. &	28.8 & 70.4	& 42.1\\
VideoStudio	& T. & 30.1	& 75.0 & 56.9 \\
\hline
SSR-Encoder	& Z. &28.6&	77.8 & 59.9\\
VideoBooth&	Z. &27.8 & 71.4	& 49.9 \\
\hline
Magic-Me &	F. &28.4 & 77.6	& 59.0 \\
DreamVideo &F. &30.2 & 72.4	& 47.2\\
\hline
Ours &F.	& \textbf{31.7}	& \textbf{80.0}	& \textbf{62.6}\\
\hline
\end{tabular*}
\caption{The quantitative results of different methods. T., Z., and F. indicate two-stage method, zero-shot method, and tuning-based method, respectively. It can be observed that our method achieves the highest performance on all metrics.}
\label{tbl:sota comparison}
\vspace{-12pt}
\end{table}

\subsection{Comparisons}
Table~\ref{tbl:sota comparison} and Figure~\ref{fig:sota comparison} summarize the quantitative and qualitative comparisons with different state-of-the-art methods. The two-stage approach relies on two separate models: a CT2I model and an I2V model. However, due to the modality gap, they often fail to integrate seamlessly, particularly when dealing with uncommon subjects. As shown in the results of VideoStudio~\cite{long2024videostudio} and TI-DB in Figure~\ref{fig:sota comparison}, the I2V model struggles to fully comprehend the structural integrity of the subject, e.g., the shape of the dog's face and ears, leading to deformation or physically implausible motion. In contrast, our approach maintains better temporal smoothness, and its consistency with the reference subject is effectively demonstrated by the CLIP-I and DINO scores. Our method outperforms the two-stage approach by 5 and 5.7 in CLIP-I and DINO scores, respectively.

\begin{figure}[ht!]
\centering
\includegraphics[width=0.5\textwidth]{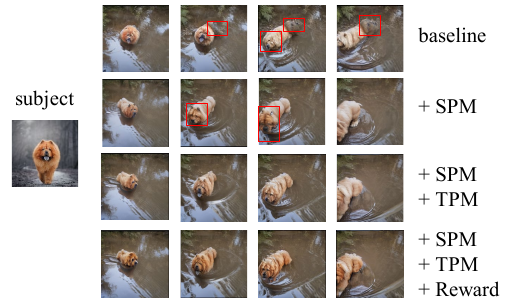} 
\caption{The visualization of the generated videos with different combination of components. It can be observed that SPM helps alleviate the structure deformation problem, TPM can enhance the texture consistency, and the reward optimization forces the color of the generated subject to approach that of the reference subject.}
\vspace{-12pt}
\label{fig:ablation study}
\end{figure}

Zero-shot approach, e.g., VideoBooth~\cite{jiang2024videobooth}, demonstrates strong temporal smoothness. However, it struggles to accurately represent the customized subject, leading to significant texture and structural discrepancies between the generated video and the reference subject. These methods rely on feature extractors trained on large-scale datasets, but inherent data biases limit their generalization to uncommon categories. In contrast, our proposed method incorporates a dedicated structure and texture propagation module, ensuring superior structural and texture consistency compared to zero-shot approaches. As a result, our method outperforms the best zero-shot approach by 2.2 and 2.7 points in CLIP-I and DINO scores, respectively.

Finally, we compare our approach with tuning-based methods, namely Magic-Me~\cite{ma2024magic} and DreamVideo~\cite{wei2024dreamvideo}. Magic-Me represents the customized subject using only a single token embedding, resulting in insufficient representational capacity and poor alignment with the reference subject. On the other hand, DreamVideo directly trains an adapter on a video model using image data, which compromises motion quality. In contrast, our method not only fine-tunes the entire model to enhance representational capacity but also autoregressively propagates structural and texture information across frames. This ensures both subject alignment and motion consistency, achieving a balanced and superior performance.

\subsection{Ablation Study}
We conducted ablation studies to further validate the effectiveness of the proposed structure and texture propagation module, as well as the test-time optimization. The results are presented in Table~\ref{tbl:ablation study} and Figure~\ref{fig:ablation study}. Our baseline method is built upon Text Inversion and DreamBooth, incorporating the motion module from AnimateDiff. As shown in the first row of Figure~\ref{fig:ablation study}, the baseline method often suffers from structural and texture inconsistencies. For instance, the dog's hindquarters exhibit an unnatural tail, and its facial features appear blurry. When the Structure Propagation Module (SPM) is introduced (second row of Figure~\ref{fig:ablation study}), the structural coherence of the dog improves significantly, leading to an increase of 4.6 and 10.5 in CLIP-I and DINO scores, respectively. However, texture inconsistencies remain—for example, the dog's face morphs into that of a different breed. After incorporating the Texture Propagation Module (TPM), the dog's facial texture becomes more consistent, further enhancing subject fidelity across frames. 

Although the consistency of structure and texture has been significantly improved at this stage, fine-grained details, such as the dog's color, still exhibit discrepancies compared to the reference subject. To address this issue, we further introduce the test-time reward optimization method. As shown in the fourth row of Figure~\ref{fig:ablation study}, after optimizing the latent representation, the dog's color distribution becomes more aligned with the reference subject, leading to further improvements in CLIP-I and DINO scores. The above analysis demonstrates that the proposed structure and texture propagation module, along with the test-time optimization method, effectively enhances the stability and consistency of customized video generation, advancing AI-driven storytelling video creation.

\begin{table}[ht!]

\centering
\begin{tabular*}{\hsize}{@{\extracolsep{\fill}}ccc|ccc}
\hline

 SPM & TPM & Reward & CLIP-T $\uparrow$	& CLIP-I $\uparrow$	& DINO $\uparrow$  \\
\hline
 & & & 29.8 & 72.2 & 49.5 \\
\checkmark &	&  & 31.1	& 76.8	& 61.0	 \\
& \checkmark & &	31.0	& 77.1	& 59.1\\
\checkmark & \checkmark &	& 31.4	&78.9 &	61.1 \\
\checkmark & \checkmark& \checkmark & \textbf{31.7}	& \textbf{80.0}	& \textbf{62.6}\\
\hline
\end{tabular*}
\caption{The quantitative results of using different combinations of components.}
\label{tbl:ablation study}
\vspace{-12pt}
\end{table}

\section{Limitation and Conclusion}
\noindent \textbf{Limitation.} Despite the improvement in maintaining subject consistency in videos, our approach still faces several limitations. First, the method struggles with personalized human subject generation. Since humans contain rich and intricate details, such as facial features and contours, the current approach finds it challenging to accurately capture and represent these fine-grained attributes. A potential solution is to incorporate prior knowledge of facial and body structures to enhance detail representation. Additionally, the length of generated videos remains relatively limited. To address this, integrating our approach with long-form video generation techniques could be a promising direction to improve the model’s capability in producing extended video sequences.

\noindent \textbf{Conclusion.} This paper primarily explores methods to enhance the consistency of customized video generation. While significant progress has been made in customized text-to-image (T2I) generation, directly integrating motion models for video generation often results in the loss of structural and texture information. To address this issue, we propose an autoregressive structure and texture propagation module. Experimental results demonstrate that our method effectively extracts the structural and texture information of the customized subject and propagates it across video frames, thereby improving overall consistency. Furthermore, we introduce a reward optimization method to refine fine-grained details, validating the feasibility of test-time optimization.
{
    \small
    \bibliographystyle{ieeenat_fullname}
    \bibliography{main}
}

\end{document}


\maketitle

\section{Supplemental Experiments}
\subsection{Implementation Details}
Our implementation is based on the open-source text-to-video (T2V) model, Animatediff~\cite{guo2023animatediff}, which is inflated from the text-to-image (T2I) generation model, Stable Diffusion~\cite{he2022latent}, by integrating the temporal motion modules. Our training, including the embedding of the special token and the fine-tuning of the T2I model weights, mainly performs at the customization stage. During this stage, we train only one token embedding for each subject with a learning rate of $10^{-3}$ for 60 steps. For the finetuning of model weights, we use a learning rate of $2 \times 10^{-5}$ for 500 steps, following the setting of \cite{pang2024attndreambooth}. In the latent enhancement stage, the wight $\lambda$ for the gradient is set to $10^2$. All experiments are conducted on a NVIDIA-A100(40G) GPU.

\subsection{Efficiency}
We analyze the training and inference time required by different methods. Both MS-Diffusion and TI-DB utilize CogVideo as their I2V model and generate 49 frames, resulting in longer inference times. Additionally, our method incorporates autoregressive structure and texture propagation along with test-time optimization, leading to the longest inference time among tuning-based approaches. However, our method achieves the best results, further validating that scaling test-time inference can lead to superior performance.

\begin{table}[ht!]

\centering
\begin{tabular*}{\hsize}{@{\extracolsep{\fill}}c|c|c|c}
\hline

  Method & Training & Inference & Num\\
\hline
 VideoStudio & - & $\sim 37$s & 16 \\
  MS-Diffusion & -  & $\sim 463$ s & 49	 \\
  TI-DB& -  & $\sim 460$ s & 49	 \\
 \hline
 VideoBooth & - & $\sim 98$s & 16\\
  SSR-Encoder & - & $\sim 12$s & 16\\
 \hline
 Dreamvideo & $\sim 26.7$ min & $\sim 76 $s & 32\\
 Magic-Me & $\sim 14.5$ min & $\sim 40$s & 16\\
 Ours & $\sim 20$ min & $\sim 120$s &16\\
 \hline
\end{tabular*}
\caption{The training and inference time costs of different methods. Num indicates the number of frames per video.}
\label{tbl:ablation study}
\vspace{-12pt}
\end{table}

\subsection{Results}
\begin{figure*}[ht!]
\centering
\includegraphics[width=1\textwidth]{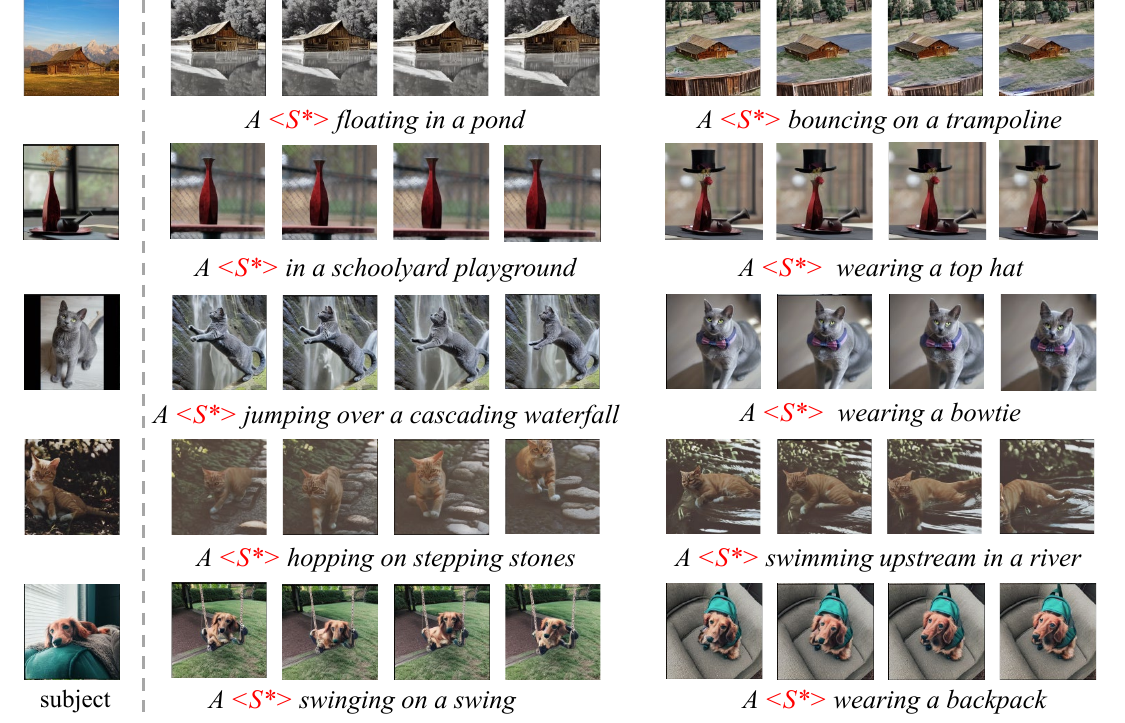} 
\caption{The visualization of qualitative results. }
\label{fig:qualitative results}
\end{figure*}

\noindent \textbf{Qualitative Results.}
We provide additional qualitative results in Figure~\ref{fig:qualitative results}, demonstrating that our method effectively maintains subject consistency for both non-living and living subjects. Moreover, our model exhibits strong adherence to prompts, as evidenced by its ability to generate modifications such as adding a hat to a vase (second row, right), attaching a bow tie to a cat (third row, right), and placing a backpack on a dog (fifth row, right). These results further validate the capability of our approach in generating diverse and personalized videos.

\begin{figure}[ht!]
\centering
\includegraphics[width=0.5\textwidth]{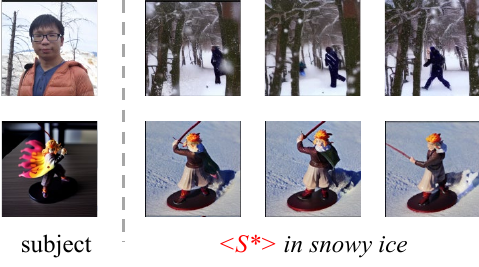} 
\caption{The visualization of two failure cases.}
\label{fig:failure cases}
\end{figure} 

\noindent \textbf{Failure Cases.} While our method effectively improves consistency for general customized subjects, it remains limited in handling human-like subjects, such as the action figure in the second row of Figure~\ref{fig:failure cases}. For instance, in the first-row example of Figure~\ref{fig:failure cases}, the generated video fails to capture facial details, instead producing a full-body representation without facial information. In the second example, the generated subject does not fully retain the characteristics of the reference subject, such as the cloak and facial details.

We hypothesize that this limitation arises due to the intricate details present in such subjects, including facial features and contours. The sequential latent representations in video generation may struggle to capture all fine-grained details, leading to loss of fidelity and blurriness. A potential solution is to incorporate more trainable parameters to enhance subject representation. For example, increasing the number of special token embeddings could allow different embeddings to encode distinct subject characteristics, thereby improving representation capacity.

Furthermore, given that these structured subjects exhibit certain patterns—such as the distribution of limbs and facial features—it may be beneficial to introduce prior knowledge of object structures. Leveraging such priors could facilitate better subject representation and improve overall generation quality.

{
    \small
    \bibliographystyle{ieeenat_fullname}
    \bibliography{main}
}